\documentclass[letterpaper, 10 pt, conference]{include/ieeeconf}  

\IEEEoverridecommandlockouts                              

\title{\LARGE \bf
Distill-then-prune: An Efficient Compression Framework for Real-time Stereo Matching Network on Edge Devices
}

\author{Baiyu pan$^{1,3}$, Jichao jiao$^{1,2}$, Jianxing pang$^{1}$, Jun cheng$^{3}$
\thanks{$^{1}$The authors are with the UBTech Robotics Corp,
Shenzhen, China
{\tt\small \{baiyu.pan,jichao.jiao,walton\}@ubtrobot.com}}%
\thanks{$^{2}$The author is with the Beijing University of Posts and Telecommunications,
Beijing, China
{\tt\small jiaojichao@bupt.edu.cn}}%
\thanks{$^{3}$The authors are with the Shenzhen Institute of Advanced Technology, Chinese Academy of Sciences,
Shenzhen, China
{\tt\small jun.cheng@siat.ac.cn}}%
}

\usepackage{cite}
\usepackage{graphicx}
\usepackage{amsmath}
\usepackage{algpseudocode}
\usepackage{makecell}
\graphicspath{ {./figs/} }
\usepackage{booktabs}
\usepackage{algorithm}
\usepackage{amssymb}
\usepackage{multicol}
\usepackage{multirow}
\usepackage{hyperref}

\newcommand{\etal}{\textit{et al.}}
\begin{document}

\maketitle
\thispagestyle{empty}
\pagestyle{empty}

\begin{abstract}



In recent years, numerous real-time stereo matching methods have been introduced, but they often lack accuracy. These methods attempt to improve accuracy by introducing new modules or integrating traditional methods. However, the improvements are only modest. In this paper, we propose a novel strategy by incorporating knowledge distillation and model pruning to overcome the inherent trade-off between speed and accuracy. As a result, we obtained a model that maintains real-time performance while delivering high accuracy on edge devices. Our proposed method involves three key steps. Firstly, we review state-of-the-art methods and design our lightweight model by removing redundant modules from those efficient models through a comparison of their contributions. Next, we leverage the efficient model as the teacher to distill knowledge into the lightweight model. Finally, we systematically prune the lightweight model to obtain the final model. Through extensive experiments conducted on two widely-used benchmarks, Sceneflow and KITTI, we perform ablation studies to analyze the effectiveness of each module and present our state-of-the-art results.







\end{abstract}

\section{Introduction}
\label{sec:intro}

Stereo matching estimates the depth information of a scene using a pair of rectified binocular images captured by cameras. The depth is determined by the disparity between the corresponding pixels in the two images.
Stereo matching is important in many fields including robot navigation\cite{bi2023application}, drone control\cite{8575255,10070843} and autonomous driving\cite{kitti2012,kitti2015}. 
However, these applications share similar requirements, including implementation on edge devices, real-time processing, and high accuracy.
In recent years, deep neural networks have achieved significant success in the field of stereo matching. In pursuit of accuracy and robustness, researchers have proposed architectures that are deeper and wider, even though this comes with a greater storage and computation time. For instance, the PSMnet \cite{chang_pyramid_2018}, which is a most commonly adopted backbone, runs $1$ fps on AGX.
This high computation requirements pose an obstacle on the feasible application of stereo matching. 

\begin{figure}
    \centering
    \includegraphics[width = 0.9\linewidth]{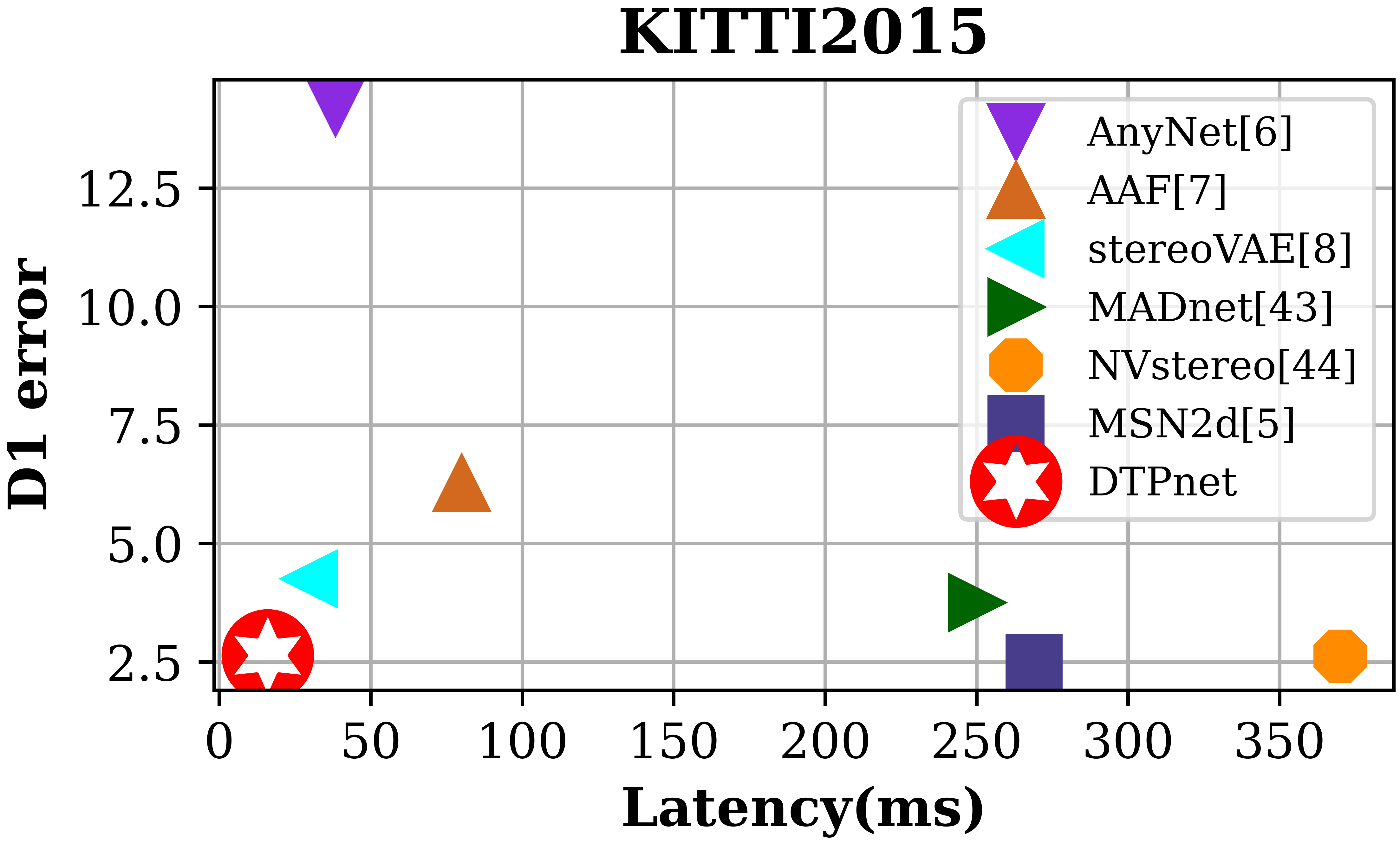}
    \caption{Latency vs. D1 error on the KITTI 2015\cite{kitti2015} validation set. The unit of Latency is millisecond/frame. Both metrics are lower the better. As shown, our \textbf{DTPnet} achieves a good balance between accuracy and speed.}
    \label{first_compare}
\end{figure}

With the goal of achieving real-time processing on edge devices (GPU/NPU), many lightweight methods have been proposed \cite{Duggal_2019_ICCV, xing2020mabnet, shamsafar_mobilestereonet_2021,wang2018anytime, Chang_2020_ACCV, StereoVAE}. Those methods can be roughly divided into two categories: multi-stage method\cite{xing2020mabnet,wang2018anytime,tankovich_hitnet_2021} and model compression method\cite{Duggal_2019_ICCV,shamsafar_mobilestereonet_2021,Chang_2020_ACCV}. 
The computational complexity of the network depends on two factors: the size of the feature map and the number of convolution kernels. Thus, multi-stage methods simultaneously decrease these two factors. For example, Wang \etal \cite{wang2018anytime} employs four coarse-to-fine stages that dynamically up-sample the feature maps as needed. 
However, this merely achieves a balance in the trade-off between accuracy and latency, without gaining any extra efficiency. 
On the other hand, the main goal of model compression is to remove the redundant part of the model to achieve better efficiency.
The Duggal \etal \cite{Duggal_2019_ICCV} proposed to progressively reduce the search space of disparity to accelerate the speed.

The typical backbone of end-to-end neural networks for stereo matching involves three main steps: feature extraction \cite{kaiming_resnet_2015,YannLeCun2015}, cost volume construction \cite{mayer_large_2016,kitti2015,YannLeCun2015}, and disparity regression module \cite{kendall_end--end_2017}. The cost volume is a 5-dimensional tensor with \textit{batch, channel, depth, height, width}. In particular, the \textit{depth} dimension is created by shifting the feature maps pixel-wise along the \textit{width}. 
The extra \textit{depth} not only increase the computation complexity and also requires 3D convolution for processing. Moreover, this introduces another challenge: the inference SDK provided by edge devices, like TensorRT\cite{tensorrt}, only provide limited support for 3D convolution operation. 
However, replacing the 3D convolution is challenging. Shamsafar \etal \cite{shamsafar_mobilestereonet_2021} has achieved promising result by compressing the \textit{channel}. More importantly, only 2D convolution is used in their network.


We conducted extensive experiments and identified three key obstacles in achieving real-time stereo matching. First, the support for operations by the device's inference SDK, which directly determines the network's feasibility on the device but is often overlooked in previous real-time-oriented studies. Second, the heavy computation, as the computational complexity of the network directly affects the processing speed. Third, low accuracy, which stands on the opposite side of the scale with computation.

In this paper, we comprehensively analyze the commonly used modules and based on our analysis, we have designed an implementation-friendly lightweight network. Moreover, we propose the distill-then-prune (\textbf{DTP}) framework to further compress and improve the network. In DTP, we leverage the knowledge distillation and model pruning, both techniques that have proven to be efficient in model compression, to enhance performance.
Overall, our main contributions are:
\begin{enumerate}
    \item A implementation-friendly lightweight network is proposed.
    \item We conduct a thorough investigation of knowledge distillation and propose a feasible scheme for stereo matching.
    \item We combine model pruning with knowledge distillation to develop the DTP framework, which is compatible with any existing stereo matching methods.
\end{enumerate}

Two benchmarks \textit{Flyingthings3D}\cite{mayer_large_2016} and \textit{KITTI}\cite{kitti2015} are leveraged for evaluation.
Extensive ablation studies are conducted to evaluate the components of our proposed networks. As shown in Fig. \ref{first_compare}, we compare against current state-of-the-art methods and backbones to demonstrate our efficiency.

\par
\par
\par
\par
\section{Related works}

\textbf{Stereo matching}, also known as disparity estimation, was regarded  as an optimization problem of locating the matching pixels between binocular images. 
The traditional methods \cite{pan2015improved,PatchMatch,SGM,Song2014MCL3DAD} can be summarized as ``search, compare, optimize'', and this behavior can still be found in the CNN-based methods\cite{LW-CNN,chang_pyramid_2018, kendall_end--end_2017,ACVnet,pan_multi-stage_2020}.
Based on \cite{GC-net}, the commonly used backbone PSMnet\cite{chang_pyramid_2018} has established the training workflow and fundamental components of the end-to-end stereo matching network.
Many methods have proposed effective modules or deeper network to improve the performance of the backbone, like atrous convolution \cite{du2019amnet}, multi-scale regression \cite{pan_multi-stage_2020,xu_aanet_2020}, volume fusion \cite{guo_group-wise_2019,chen2021cost}.


Besides the aforementioned real-time-oriented methods, there are other multi-stage methods. HITnet\cite{tankovich_hitnet_2021} has adopted a structure similar to that of \cite{wang2018anytime}, but with the addition of geometric warping to obtain multi-resolution results. 
Xing \etal\cite{xing2020mabnet} has proposed adjust multi-branch module which combines depth-wise convolution to reduce the number of channels.
The mobilestereonet's\cite{shamsafar_mobilestereonet_2021} main contribution is that they have proposed to reform the cost volume with convolution and use the 2D convolution only for the disparity regression.
However, the final FLOPs and latency are still too larger and far from real-time.

\textbf{Knowledge distillation}\cite{knwoledge_vanilla,turc2020wellread,teacherKnowledge,liu2022learning,DWARF,zhang2021improve} is an effective tool for improving the performance of the compressed neural network. 
Hinton \etal\cite{knwoledge_vanilla} proposed the vanilla logits-based knowledge distillation framework. In this framework, the student network learns from the logits of the teacher network, using a temperature hyperparameter to control the difficulty of learning. While the logits-based method is suitable for a wide range of tasks and models, its limitation lies in the fact that the student cannot learn the internal representations of the teacher.
Zhang et al. \cite{zhang2021improve} proposed distilling the network by minimizing the distance of extracted features between the teacher and student networks. The main advantage of feature-based knowledge distillation is that the student can learn more informative and robust representations from the teacher. However, feature-based knowledge is difficult to train \cite{liu2022learning} and requires a delicate training protocol to acquire meaningful knowledge.
In this paper, we have compared both schemes, and the results show that only the logits-based approach \cite{knwoledge_vanilla} is feasible in the task of stereo matching.

\textbf{Model pruning} is an efficient tool for compress the neural network. The methods of model pruning have two main categories: non-structure pruning\cite{han2015deep_compression,han2015learning,frankle_lottery_2019} and structure pruning\cite{DBLP:journals/corr/abs-2001-08565,luo_thinet_2017,liu2022learning,fang2023depgraph}. The non-structure methods are implementation-unfriendly for edge devices due to their sparse structure. The structure method prunes certain filters of a layer based on their importance. Luo \etal \cite{luo_thinet_2017} rank and prune the filters based on their activation. In this paper, we follow the method \cite{fang2023depgraph} to prune the network and obtain a compact network.

\section{Proposed lightweight network}
We designed our network based on two goals, \textit{lightweight} and \textit{implementation-friendly}. 
To achieve \textit{lightweight}, we compare the commonly used modules\cite{kaiming_resnet_2015,StackedHourglass,chang_pyramid_2018,wang2018anytime} and remove the redundant ones. 
And to make it \textit{implementation-friendly}, we have made three main changes: 1. We have replaced all the 3D convolution kernels with 2D convolution kernels. 2. The iterative cost volume construction is replaced with a channel-to-disparity module. 3. The commonly used trilinear interpolation after softmax layer is replaced by a bilinear interpolation in the inference stage.
\subsection[short]{Architecture}
In our network, three modules are included, \textit{Feature extraction network}, \textit{Cost volume construction} and \textit{Disparity regression}.
The feature extraction network is a siamese network that shared the same weight, and process the input binocular image pairs parallel. After obtaining the features $f_{l}$ and $f_{r}$ from the left and right images, the cost volume construction is responsible for integrating them into a cost volume.
Finally, the disparity regression module is responsible for predicting the disparity based on the cost volume.

\begin{table}[h]
\centering
\caption{The parameters and FLOPs of each module.}
\begin{tabular}{lcc}
\toprule
Moudle& Params$_{(M)}$&FLOPs$_{(G)}$\\
\midrule
$Feature\ extraction$       & 0.10 & 0.72 \\\midrule
$Cost\ volume\ construction$    & 0.03 & 0.11 \\\midrule
$Disparity\ regression$    & 0.51 & 5.42  \\\midrule
\textbf{Total}  & 0.64 & 6.25 \\\bottomrule
\end{tabular}
\label{table_architecture}
\end{table}

In Table \ref{table_architecture}, we present the details of the proposed lightweight network. We conducted extensive ablation studies to compare the commonly used architectures and number of layers. By comparing the gains and losses in terms of parameters and accuracy, we removed the redundant modules and arrived at our final architecture. More details can be found in Section \ref{Experimental results}.

\textbf{Feature Extraction Network}: Our network leverages the feature pyramid with two residual blocks to extract and concatenate multi-scale features. The outputs $\{f_{l},f_{r} \in \mathbb{R}: |f|=(B\times C\times \frac{H}{4}\times \frac{W}{4})\}$ are the features of left and right image respectively. 
\par
\textbf{Cost Volume Construction}: Shamsafar \etal\cite{shamsafar_mobilestereonet_2021} proposed to iteratively construct the cost volume while simultaneously applying channel compression through convolutional layers. However, the total number of floating-point operations will increase cumulatively with each iteration. Therefore, we proposed a channel-to-disparity module where we directly compress the channel and guide the convolutions in learning the mapping from channel to disparity.

The module is composed by $3$ layers of conventional layer:
\begin{equation}
  y = \mathcal{F}(concat(f_{l},f_{r}),\{W_{i}\}).
\end{equation}
The $f_{l},f_{r}$ are concatenated first, and take as input of $\mathcal{F}$. The function $\mathcal{F} = W_{i}\sigma(W_{i-1}\sigma(...W_{0}(\cdot)...))$ in which $\sigma$ denotes the activation layer and batch normalization layer, and $i=3$.

\textbf{Disparity Regression Module} includes three parts, the stacked hourglass module\cite{StackedHourglass,chang_pyramid_2018}, upsampling layer and softmax layer. 

\begin{figure}[h]
  \centering
  \includegraphics[width=0.9\linewidth]{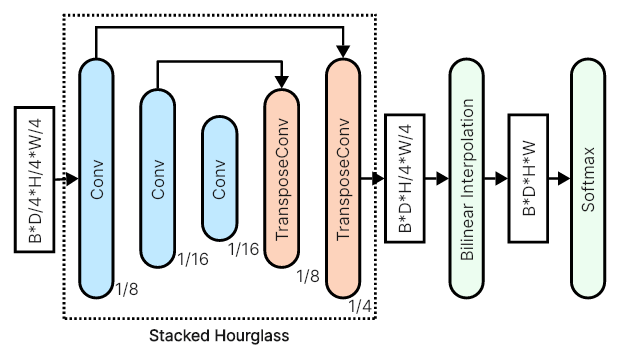}
  \caption{Disparity regression module. The stacked hourglass module is composed by convolutions and transpose convolutions. The arrow line above denotes the skip connection.}
  \label{StackedHourglass}
\end{figure}

As shown in Fig. \ref{StackedHourglass}, the encoder-decoder structure of the stacked hourglass provides a larger receptive field for feature maps, enabling it to capture contextual and semantic information. 
Moreover, as shown in Table \ref{table_architecture}, the stacked hourglass module has the most parameters and FLOPs compared to others, but it also results in the most significant improvement in the outcomes. Further details can be found in Table \ref{Inter-comparison}.

The upsampling layer is used to upsample the feature to $\{p\in \mathbb{R}^{3}: |p| = (B\times d_{max}\times H\times W)\}$ where the hyperparameter $d_{max}$ is set to 192. 


After softmax, a soft-argmax (Equ. \ref{softargmax}) is adopted for transform the probabilities to  the predicted disparity result $\mathcal{D}$,
\begin{equation}
\mathcal{D} = \sum_{i=0}^{d_{max}}d_{i}*\frac{exp(p_{i})}{exp(\sum_{j}^{d_{max}}p_{j})}.
\label{softargmax}
\end{equation}
In the typical training workflow of disparity estimation network, the $smooth_{l1}$ loss is employed \cite{chang_pyramid_2018}.
However, in our approach, we train the network using our proposed DTP framework, and the training details are provided in Section \ref{Distill-Then-Prune framework}.

\subsection{Adaptation for implementation}

During implementation, we discovered that several operations, including 3D convolutions, slicing, iteratively applied convolutions, and trilinear interpolation, are not supported by the hardware. As a result, we implemented the aforementioned changes to address these issues while still achieving competitive results.

The 3D convolution is employed because of the 5D cost volume. In MobileStereoNet\cite{shamsafar_mobilestereonet_2021}, they propose an iterative convolution approach to compress the channel of the cost volume and replace the 3D convolution with 2D convolution.

\begin{equation}
\begin{matrix}
C(c,d,x,y)=g_{c}(f_{l}(.,x,y),f_{y}(.,x-d,y))
\\ 
\Downarrow 
\\
C(1,d,x,y)=g_{c=1}(f_{l}(.,x,y),f_{y}(.,x-d,y)),
\end{matrix}
\end{equation}
where $c$ represents the channel, and $g_{c}$ denotes the convolution layers with an output channel $c$. During each iteration, MobileStereoNet\cite{shamsafar_mobilestereonet_2021} compressed the channel to $1$ and then squeezed it afterward to obtain the 4D cost volume.
Inspired by their work, we have designed our channel-to-disparity module, which completely eliminates the need for iterations and slice operations.

The remaining challenge is the trilinear interpolation. It is employed to upsample the $D, H, W$ of the feature simultaneously.
We have introduced an equivalent two-step operation, where a convolutional layer is utilized to map the channel from $D/4$ to $D$, followed by bilinear interpolation to upsample the feature map's $H$ and $W$ dimensions.

\section{Distill-Then-Prune framework}
\label{Distill-Then-Prune framework}
As the name of our proposed Distill-Then-Prune (DTP) framework indicates, it includes two main parts: \textit{knowledge distillation} and \textit{model pruning}. In Algorithm \ref{alg_DTP}, we provide details of our framework. 
\begin{algorithm}[h]
  \caption{Distill-Then-Prune(DTP) training framework}    
  \label{alg_DTP}
\begin{algorithmic}[1]
  \renewcommand{\algorithmicrequire}{\textbf{Input:}}
  \renewcommand{\algorithmicensure}{\textbf{Output:}}
  \Require Binocular images $I_{l},I_{r}$;  Max training step $M$; Learning rate $\gamma$; Parameter of teacher model $\theta_{T}$; Pruning rate $r$ and step $E$;
  \Ensure  Parameter of student model $\theta_{S}$;
  \State Initialize parameters of model: $\theta_{S}$;
  \While{$m<M$} \Comment{Knowledge Distillation}
  \State $\theta_{s} \leftarrow \theta_{s}-\gamma \frac{\partial \mathcal{L}oss(\theta_{s},\theta_{t})}{\partial \theta_{s} }$
  \EndWhile
  \While{$e<E$} \Comment{Pruning}
  \State $\theta_{S}^{e} \leftarrow prune(\theta_{S}^{e-1},r)$
  \While{$m<M$} \Comment{Finetune by Distillation}
  \State $\theta^{e}_{S} \leftarrow \theta_{S}^{e}-\gamma \frac{\partial \mathcal{L}oss_(\theta_{S}^{e},\theta_{T} )}{\partial \theta_{S}^{e} }$
  \EndWhile
  \EndWhile
\end{algorithmic}
\end{algorithm}

It's worth emphasizing that our DTP framework can be applied to any stereo matching network, without being limited to training lightweight networks. After completing the DTP, $r*E$ percent of the model's parameters will be pruned.

\subsection{Knowledge distillation}

Our knowledge distillation has a teacher model and a student model. We train the student model to learn from the teacher's logits directly. 
Our loss function is defined as,
\begin{equation}
\mathcal{L}oss(p,q) = \sum_{i=0}^{d_{max}}|\frac{exp(p_{i}/t)}{exp(\sum_{j}^{d_{max}}(p_j/t))}  -  \frac{exp(q_{i}/t)}{exp(\sum_{j}^{d_{max}}(q_j/t))}|_{1}.
\label{kdloss}
\end{equation}
The temperature $t$ is used to control the  learning difficulties. The $t$ grows from $0.5$ to $1.0$ as the epoch goes. 

Additionally, we have conducted comparisons with other settings, such as combined training with ground truth, a comparison between KL loss and L1 loss. Further details are available in Section \ref{Experimental results}.

\subsection{Model pruning}

We have adopted Depgraph \cite{fang2023depgraph} to conduct structural pruning. First, a dependency graph of the trained student network is constructed. The dependency graph groups the coupled convolutions between paired layers. 

A simple L2-norm-based scheme is adopted to calculate the importance of convolution kernels, $w=\{w_{1},w_{2},...,w_{j}\}$. 
The importance of kernels is defined as $I(w)=\{\|w_{i} \|_{2} : w_{i}\in w\}$. Based on the ranking of $I(w)$, $r\%$ of the kernels is marked.
Subsequently, adjacent layers with the same pruning scheme are grouped together. Then, the aggregated importance of a group $I(g)$ is calculated by, 
\begin{equation}
  I(g) = \sum_{w_{i}\in g}\|w_{i} \|_{2}.
\end{equation}
Finally, we prune the entire group based on the ranking of their aggregated importance, $I(g)$.

\section{Experimental results}
\label{Experimental results}
In this section, we will refer our network as \textbf{DTPnet}.
We conducted experiments on two datasets:
\textbf{SceneFlow}\cite{mayer_large_2016} is a large scale of synthetic stereo dataset which contains more than $35$k training pairs and $4.3$k testing pairs with resolution 960x540. 
\textbf{KITTI}\cite{kitti2012,kitti2015} includes several driving-scene related subsets and challenges. We use \textit{KITTI2015}\cite{kitti2015} for train and test. It contains $200$ pairs for training and $200$ pairs for testing; with resolution $1242\times 375$.

\textbf{Metrics}: End-point error (EPE)\cite{mayer_large_2016} is commonly used to optical flow evaluation.
D1 error \cite{kitti2015} are used for KITTI, D1 calculates the percentage of error pixels to the whole image. Pixel with EPE larger than $3$ will be considered as error.
\subsection{Implementation details}
Our method was implemented using the PyTorch. 
Our hand-trained PSMnet\cite{chang_pyramid_2018} with EPE $0.69$ is adopted as the teacher network. The training of the teacher is following the standard training protocol.

In DTP, we train the student network with AdamW\cite{Adamw}, with $\beta_{1}=0.9,\beta_{2}=0.999$, learning rate $=1e^{-3}$ and weight decay rate $=1e^{-2}$. The training protocol is to train on SceneFlow for $20$ epochs. 
Then, we finetune on KITTI for $300$ epochs with $1e^{-3}$, then decay the weight to $1e^{-4}$ for another $300$ epochs. We set prune rate $r=0.1$ and step $E=5$. At the end of the prune, 50\% parameters of the model are pruned. At each iteration, we finetuning the pruned network for $5$ epochs on Sceneflow, and $100$ epochs on KITTI.

\subsection[short]{Ablation study}
We have conducted comprehensive ablation studies to compare the architecture of our model and DTP framework. Firstly, we compare the commonly used modules of the backbone and illustrate the efficiency of our network. Secondly, we demonstrate the necessity of knowledge distillation and compared different settings.

\subsubsection{\textbf{Comparison of modules}}
As aforementioned, three modules are commonly used in disparity estimation\cite{chang_pyramid_2018,shamsafar_mobilestereonet_2021}. And these modules have various different settings internally, such as the repeated times of the layers.
Therefore, we conducted the experiments from two perspectives: the first being the \textit{inter-comparison}, in which we compared the influence on results of the internal layer settings of each module. The second is the \textit{intra-comparison}, where we compared the influence on results between the different modules. 

\textit{a. \textbf{Inter-comparison}}: In Table \ref{Inter-comparison}, we compared three settings $\{1,2,3\}$. The setting $1$ is proposed based on \cite{shamsafar_mobilestereonet_2021}. 
The setting $3$ is our adopted network structure for \textbf{DTPnet}.

\begin{table}[h]
\centering
\caption{Inter-comparison: The parameters and FLOPs of modules}
\begin{tabular}{lcc|cc|cc}
\toprule
& \multicolumn{2}{c}{Setting1} &\multicolumn{2}{c}{Setting2}  & \multicolumn{2}{c}{Setting3 (DTP)}\\\midrule
\multirow{2}{*}{Module}& {\tiny Params}& {\tiny FLOPs} &   {\tiny Params}& {\tiny FLOPs}&   {\tiny Params}& {\tiny FLOPs}\\
&(M)&(G)&(M)&(G)&(M)&(G)  \\
\midrule
$Feature_{1}$      & 0.01 & 0.72 & 0.01 & 0.72 & 0.01 & 0.72   \\
$Feature_{2}$      & 0.16 & 3.17 & 0.09 & 0.50 & 0.09 & 0.50   \\
$Feature_{3}$      & 0.10 & 2.21 & 0.10 & 2.21 & $/$ & $/ $   \\
$Feature_{4}$      & 0.11 & 2.13 & 0.11 & 2.13 & $/$ & $/ $   \\\midrule
$CostVolume$       & 0.13 & 7.16 & 0.13 & 7.16 & 0.03 & 0.11   \\\midrule
$Hourglass_{1}$    & 0.51 & 2.67 & 0.51 & 2.67 & 0.51 & 2.67   \\
$Hourglass_{2}$    & 0.51 & 2.67 & $/ $ & $/ $ & $/$ & $/ $   \\
$Hourglass_{3}$    & 0.51 & 2.67 & $/ $ & $/ $ & $/$ & $/ $   \\\midrule
\textbf{Total}     & 2.04 & 23.4 & 0.95 & 15.4 & 0.64 & 4.00\\\midrule
\textbf{EPE}     & \multicolumn{2}{c|}{1.27}  & \multicolumn{2}{c|}{1.48\textbf{\tiny(+0.21)}} &   \multicolumn{2}{c}{1.73\textbf{\tiny(+0.46)}}\\
\textbf{Latency}$^{*}$   & \multicolumn{2}{c|}{104 ms}  & \multicolumn{2}{c|}{44\textbf{\tiny(-60)} ms} &   \multicolumn{2}{c}{\textbf{16{\tiny(-88)} ms}} \\\bottomrule
\multicolumn{7}{l}{\footnotesize  ``$/ $'' denotes the layer is not adopted.}\\
\multicolumn{7}{l}{\footnotesize  $*$ Latency is single frame inference time on Nvidia Jetson AGX.}\\
\end{tabular}
\label{Inter-comparison}
\end{table}
\begin{table*}[t]
  \centering
  \caption{The quantitative comparison with other lightweight methods on KITTI2015\cite{kitti2015}}
  \begin{tabular}{l|ccc|ccc|cc}
    \toprule
  & ALL-D1$_{bg}$     & ALL-D1$_{fg}$     & ALL-D1$_{all}$    &  NOC-D1$_{bg}$      & NOC-D1$_{fg}$     & NOC-D1$_{all}$ & FLOPs(G) & Latency{\tiny \textbf{(ms)}}\\ \midrule
  AnyNet\cite{wang2018anytime}  & 14.2 \% & 12.1 \% & 8.51 \%    &   /   & /& / &0.04 &38.4\textbf{\tiny(AGX)}\\
  AAF\cite{Chang_2020_ACCV}    & 6.27 \%   & 13.9 \%  & 7.54 \%   &    5.96 \%	& 13.01 \%	&7.12 \% &0.02 & 80\textbf{\tiny(Tx2)}\\
  StereoNet\cite{khamis_stereonet_2018} & 4.30 \%    & 7.45 \%   & 4.83 \%  &  /     & /&/ &0.36 &1000+\textbf{\tiny(Tx2)}\\
  StereoVAE\cite{StereoVAE}    & 4.25 \%   & 10.1 \%  & 5.23 \%   &    3.88\%	&8.94 \%&	4.71 \% &/ &29.8\textbf{\tiny(AGX)} \\    
  MADnet\cite{MADnet}      & 3.75 \%   & 9.20 \%   & 4.66 \%           &  3.45 \%&	8.41 \%&	4.27 \%	&3.82& 250\textbf{\tiny(Tx2)}\\
  MABnet\cite{xing2020mabnet}      & 3.04 \%   & 8.07 \%   & 3.88 \%     &    2.80 \%& 	7.28 \%	&3.54 \%  &0.04   &/ \\
  DWARF\cite{DWARF} & 3.20 \%   & 3.94 \%   & 3.33 \%       &   2.95 \%	&3.66 \%	&3.07 \%    &19.6    & 1000+\textbf{\tiny(Tx2)}\\ 
  \href{https://www.cvlibs.net/datasets/kitti/eval_scene_flow_detail.php?benchmark=stereo&result=77e2ffe05f35444bc1d61761468c49529f5fe99f}{\textbf{DTPnet}(ours)}
  & \underline{2.64 \%}   & \underline{6.47 \%}  & \underline{3.28 \%} &  \underline{2.46 \%}&	\underline{5.61 \%}&	\underline{2.98 \%}  &\underline{\textbf{0.63}} & \textbf{16.3}\textbf{\tiny(AGX)}\\
  NVStereo\cite{NVstereo} & 2.62 \%   & 5.69 \%   & 3.13 \% &      \textbf{2.03} \%&	4.41 \%&	\textbf{2.42} \%     &3.10  & 370\textbf{\tiny(Tx2)}  \\
  AdaStereo\cite{adastereo}    &2.59 \%	&5.55 \%&	3.08 \%  &2.39 \% &5.06 \% &2.83 \% &9.37 &/\\		
  MSN2d\cite{shamsafar_mobilestereonet_2021}    & 2.49 \%   & {4.53} \%   & 2.83 \% & 2.29 \%	&3.81 \%&	2.54 \% &2.23 & 269\textbf{\tiny(AGX)}\\
  DeepPruner\cite{Duggal2019ICCV}    & 2.32 \%   & \textbf{3.91} \%   & \textbf{2.59} \% & 2.13 \%	&\textbf{3.43} \%&	\textbf{2.35} \% &103.6 &/\\\bottomrule
  \multicolumn{8}{l}{ * The latency is a single frame inference on Nvidia Tx2 or AGX}
\end{tabular}
\label{table_kitti2015}
\end{table*}

The $Feature_{*}$ denotes the feature extraction modules, each includes $4$ residual blocks\cite{chang_pyramid_2018}. The disparity regression module includes $3$ identical stacked hourglass blocks, $Hourglass_{*}$. We reduced the repetition of each block and compared the results. And leading to the Setting 3 as the one that best suited our objectives.

\textit{b. \textbf{Intra-comparison}}: The SPP\cite{SPP_He} module was adopted in \cite{chang_pyramid_2018,shamsafar_mobilestereonet_2021} to aggregate the cross spatial features from the feature extraction pyramid. 
In our network, we have reduced the pyramid to two layers. Therefore, the SPP module is redundant and does not improve the result.

\begin{table}[h]
\centering
\caption{Intra-comparison: The parameters and FLOPs of modules}
\begin{tabular}{l|ccc}
\toprule
\multirow{2}{*}{Module}&   {\tiny Params}& {\tiny FLOPs}&\multirow{2}{*}{EPE}\\
&(M)&(G)  \\
\midrule 
$Feature$       & 0.10 & 1.22  &  1.76 \\
$\ +SPP$\cite{SPP_He}      & 0.11\textbf{\tiny(+0.01)}  & 1.62\textbf{\tiny(+0.40)}   & 1.75\textbf{\tiny(-0.01)}  \\\midrule
$CostVolume$\cite{shamsafar_mobilestereonet_2021}  & 0.13 & 7.16 & 1.57  \\
$CostVolume_{our}$      & 0.03\textbf{\tiny(-0.1)}  & 0.11\textbf{\tiny(-7.05)}  & 1.73\textbf{\tiny(+0.16)}  \\\midrule
$Setting3$          & 0.63 & 6.30  & 1.73 \\
$\ - hourglass$     & 0.13\textbf{\tiny(-0.5)} & 3.63\textbf{\tiny(-2.67)}  & 5.04\textbf{\tiny(+3.31)} \\
\bottomrule
\end{tabular}
\label{Intra-comparison}
\end{table}

The $CostVolume_{our}$ denotes our channel-to-disparity module. 
The iterative channel compression module\cite{shamsafar_mobilestereonet_2021} achieves a result that is $0.16$ higher than ours. However, it also incurs an additional $7$G FLOPs compared to ours. Therefore, we are satisfied with this trade-off.
Additionally, we want to highlight that, based on our observations, the stacked hourglass module contributes the most to the overall results. Removing it results in a reduction of nearly $3$G FLOPs, as shown in Table \ref{Intra-comparison}, but it also leads to a significant drop in accuracy.

\subsubsection{\textbf{Comparison of knowledge distillation}}
We compare the training protocol of knowledge distillation from two aspects, the supervision signal, and training loss.


\textit{a. \textbf{Supervision signal}}:
In the vanilla knowledge distillation\cite{knwoledge_vanilla}, the student network is learned from the teacher and ground truth simultaneously. This training protocol was designed for the object detection task, where the ground truth is sparse. However, in the task of disparity estimation, the ground truth is a dense disparity map, more similar to the softmax output. Therefore, we argue that supervision by the teacher alone is sufficient. Thus, we designed three pairs of supervision signals: $1$. Ground truth only, $2$. Ground truth and knowledge from the teacher, $3$. Knowledge from the teacher only. As demonstrated in Table. \ref{table_targets}, the results confirm our hypothesis. 

\begin{table}[h]
\centering
\caption{Ablation study on supervision signal}
\begin{tabular}{l|c}
\toprule
Signal&  EPE \\
\midrule 
Ground truth only        & 1.73 \\
Knowledge + Ground truth & 1.61 \\
Knowledge only           & \textbf{1.56} \\\bottomrule
\end{tabular}
\label{table_targets}
\end{table}

\textit{b. \textbf{Comparison of training loss}}:
\textit{Kullback-Leibler} divergence is commonly used in knowledge distillation method\cite{knwoledge_vanilla,zhang2021improve,teacherKnowledge}. However, based on our assumption that the supervision signal from the teacher's knowledge is similar to the ground truth, and considering that the L1-norm is a standard choice in training disparity estimation networks, we assert that the L1-norm is a viable alternative. As demonstrated in Table \ref{compare_loss}, models trained with the L1-norm loss have achieved superior results.

\begin{table}[h]
\centering
\caption{Comparison of training loss}
\begin{tabular}{l|c|c}
\toprule
       &KL divergence& L1-norm\\\midrule 
EPE    &1.86  & \textbf{1.78} \\\bottomrule
\end{tabular}
\label{compare_loss}
\end{table}

\subsection{Quantitative comparison results}

\begin{figure*}[ht]
    \rotatebox{90}{Input}
    \includegraphics[width=0.3\linewidth,height=0.09\linewidth]{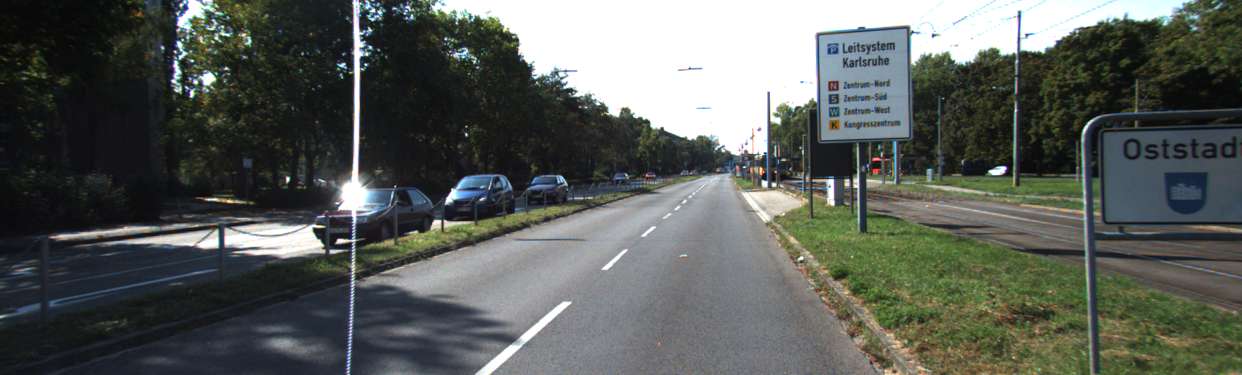} 
    \includegraphics[width=0.3\linewidth,height=0.09\linewidth]{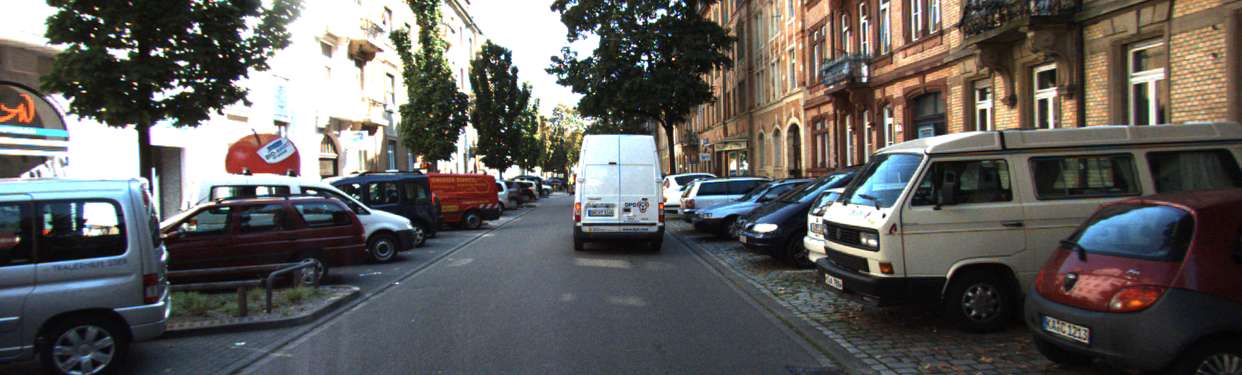}
    \includegraphics[width=0.3\linewidth,height=0.09\linewidth]{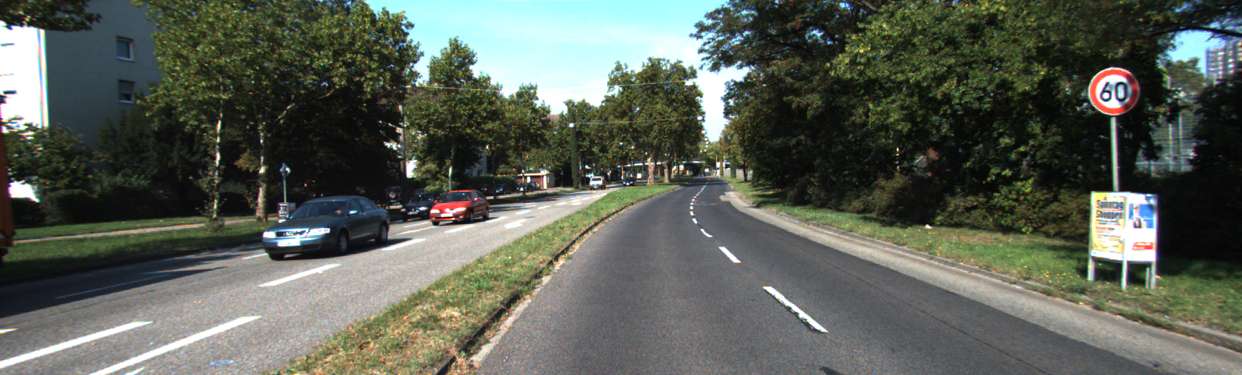}\\
    \rotatebox{90}{Aafs\cite{Chang_2020_ACCV} }
    \includegraphics[width=0.3\linewidth,height=0.09\linewidth]{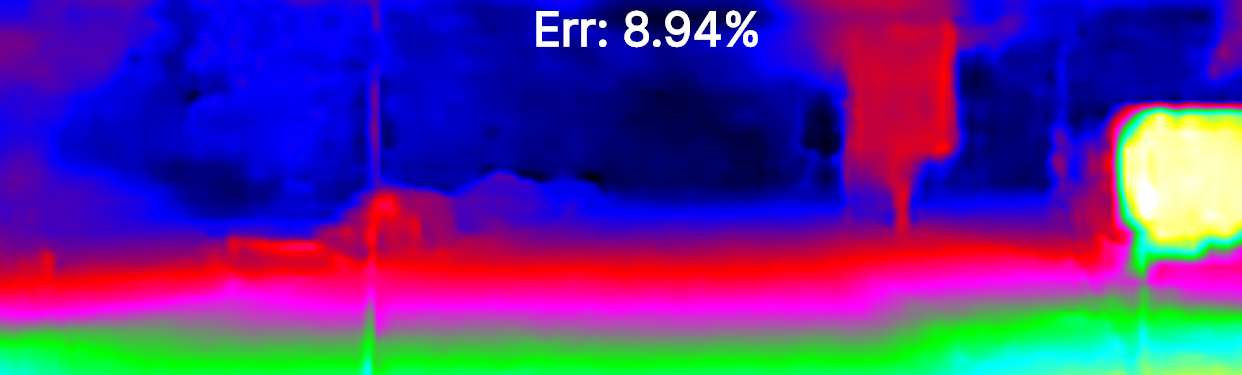} 
    \includegraphics[width=0.3\linewidth,height=0.09\linewidth]{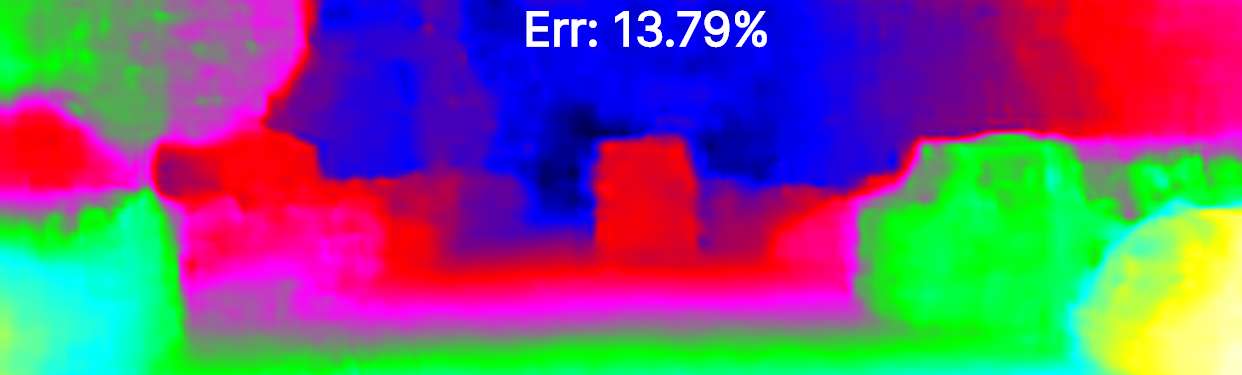}
    \includegraphics[width=0.3\linewidth,height=0.09\linewidth]{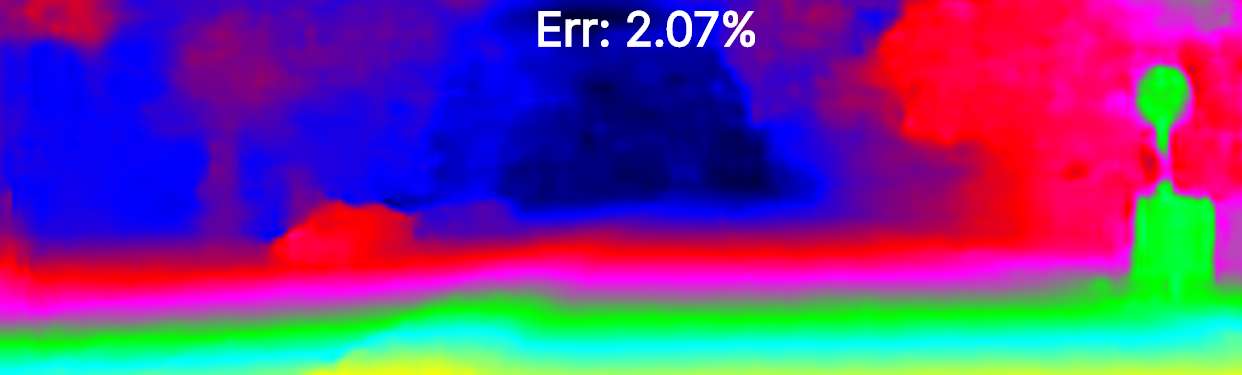}\\
    \rotatebox{90}{Aafs\cite{Chang_2020_ACCV} }
    \includegraphics[width=0.3\linewidth,height=0.09\linewidth]{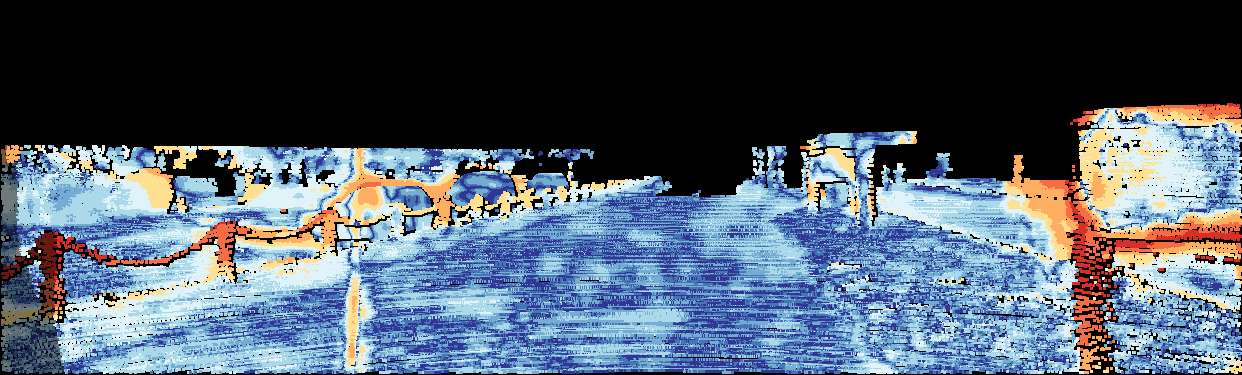} 
    \includegraphics[width=0.3\linewidth,height=0.09\linewidth]{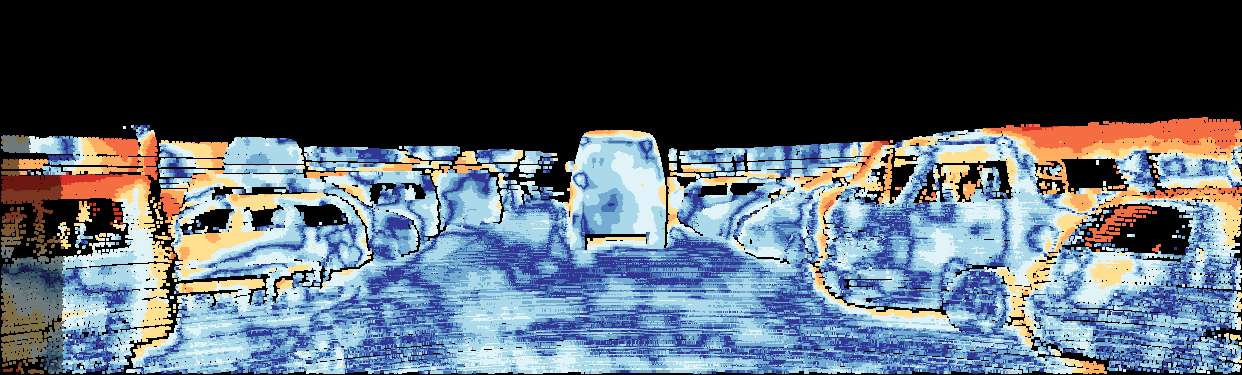}
    \includegraphics[width=0.3\linewidth,height=0.09\linewidth]{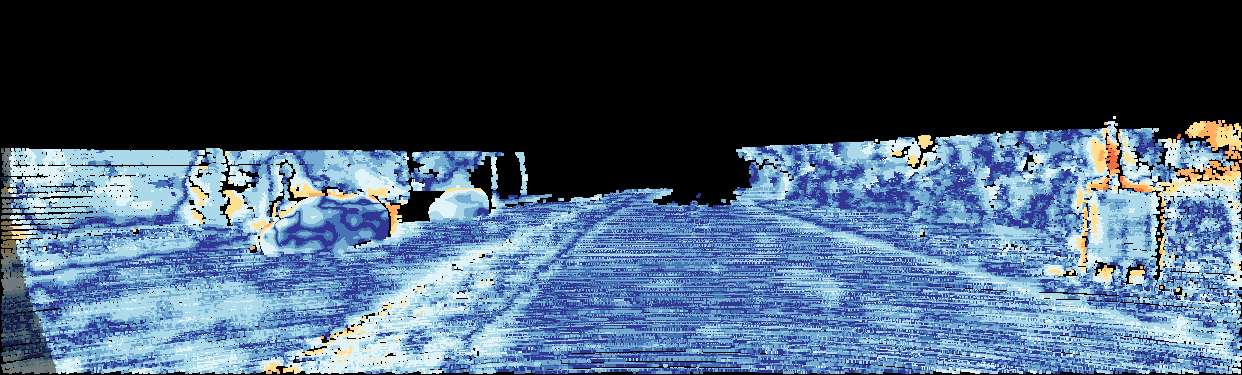}\\
    \rotatebox{90}{VAE\cite{StereoVAE}}
    \includegraphics[width=0.3\linewidth,height=0.09\linewidth]{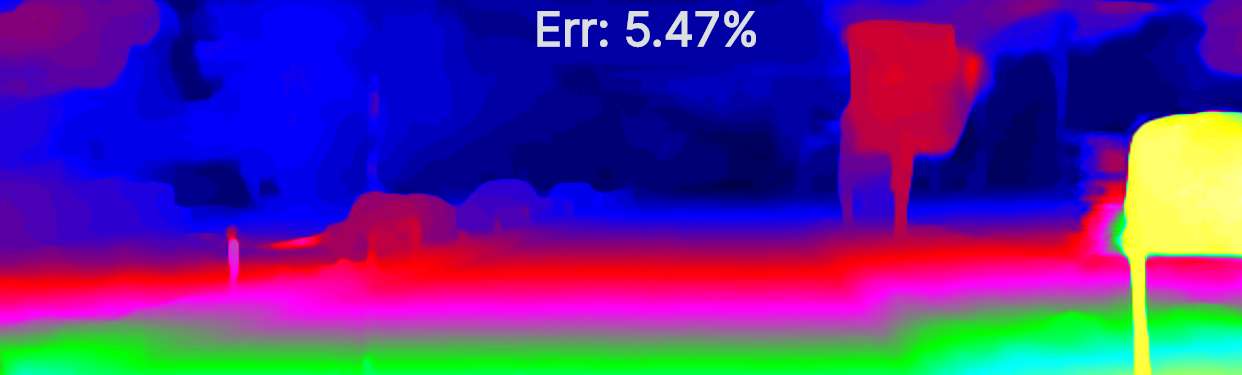} 
    \includegraphics[width=0.3\linewidth,height=0.09\linewidth]{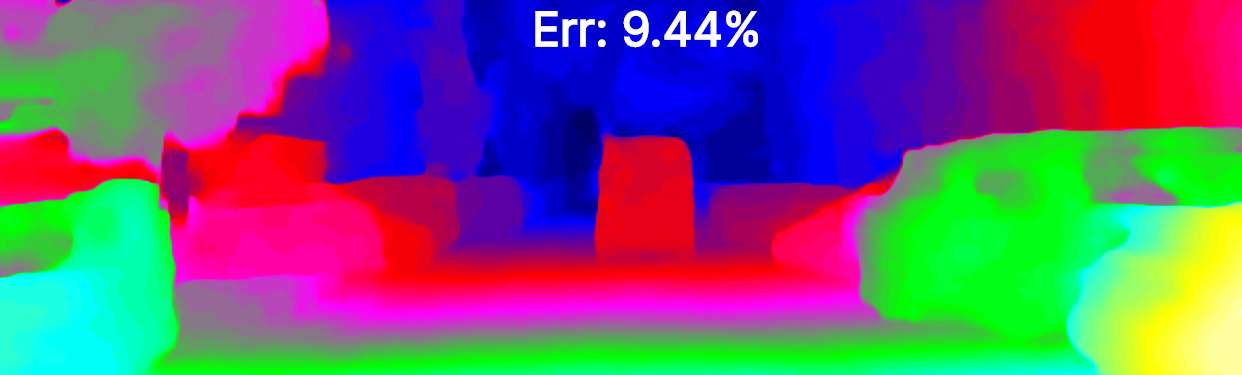}
    \includegraphics[width=0.3\linewidth,height=0.09\linewidth]{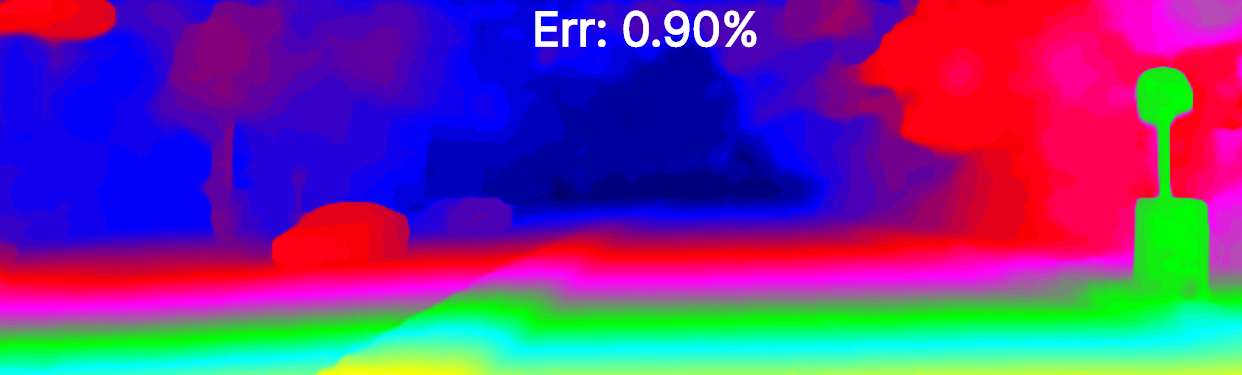}\\
    \rotatebox{90}{VAE\cite{StereoVAE}}
    \includegraphics[width=0.3\linewidth,height=0.09\linewidth]{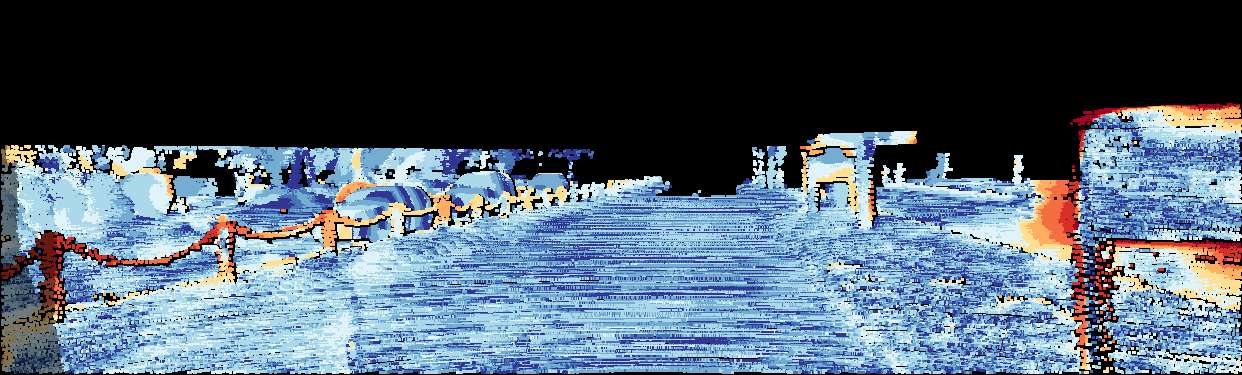} 
    \includegraphics[width=0.3\linewidth,height=0.09\linewidth]{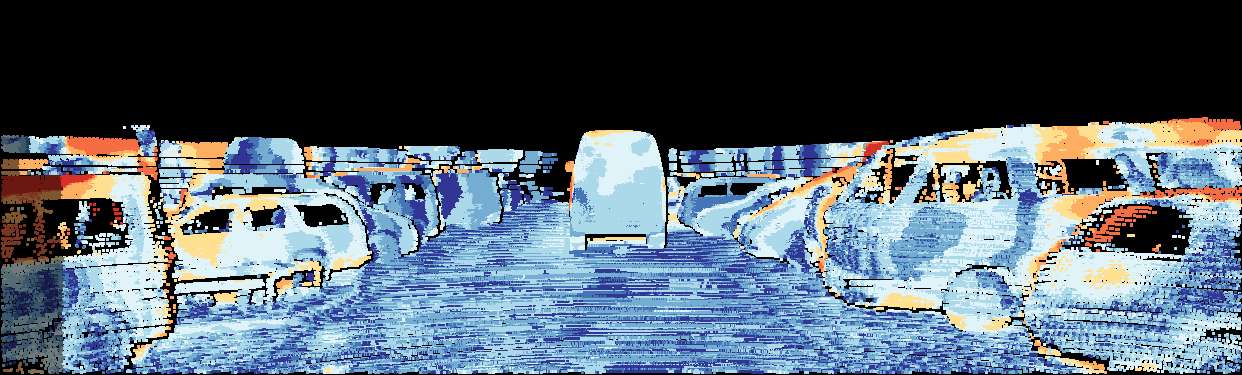}
    \includegraphics[width=0.3\linewidth,height=0.09\linewidth]{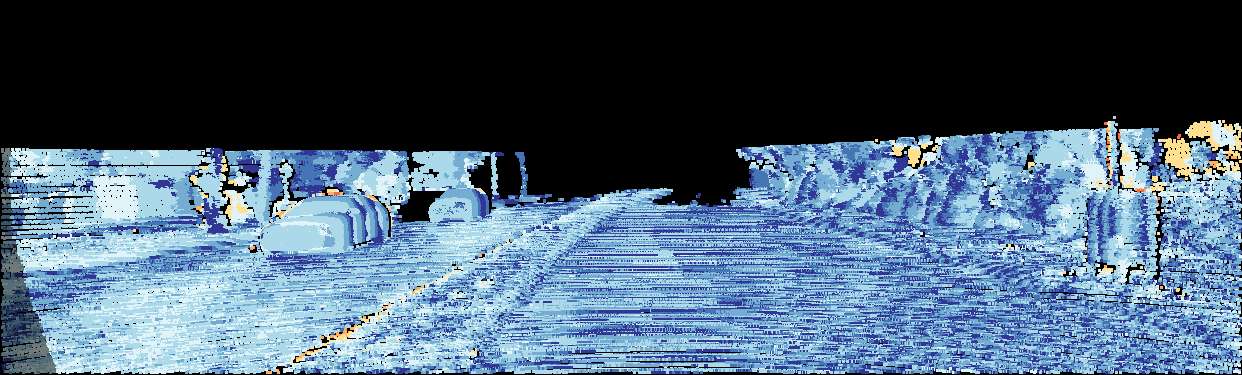}\\
    \rotatebox{90}{Ours}
    \includegraphics[width=0.3\linewidth,height=0.09\linewidth]{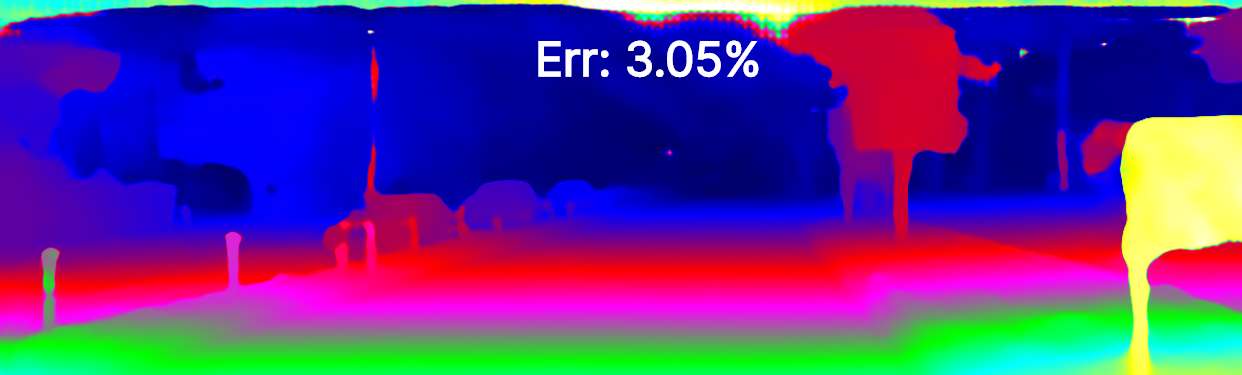} 
    \includegraphics[width=0.3\linewidth,height=0.09\linewidth]{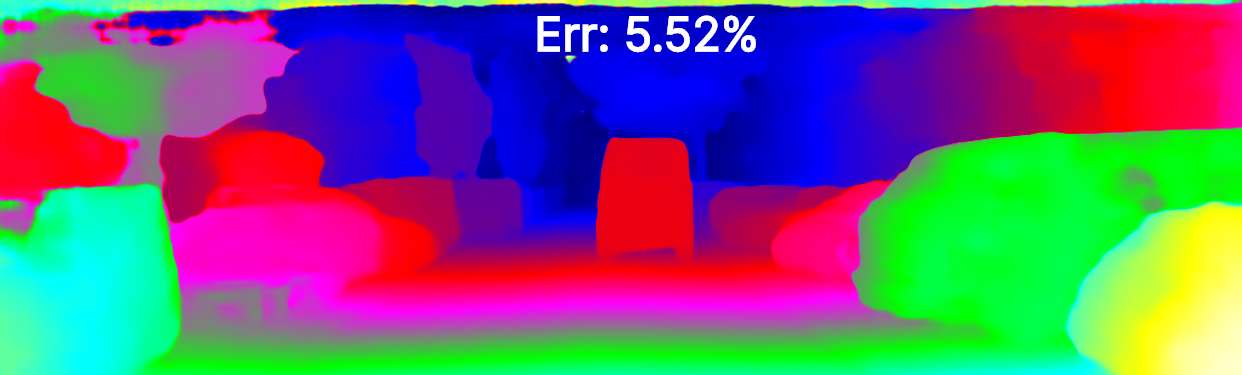}
    \includegraphics[width=0.3\linewidth,height=0.09\linewidth]{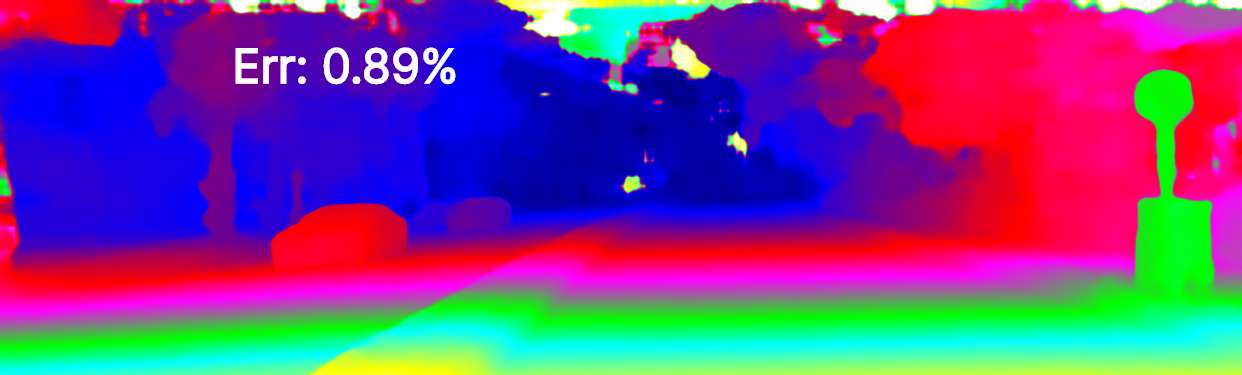}\\
    \rotatebox{90}{Ours}
    \includegraphics[width=0.3\linewidth,height=0.09\linewidth]{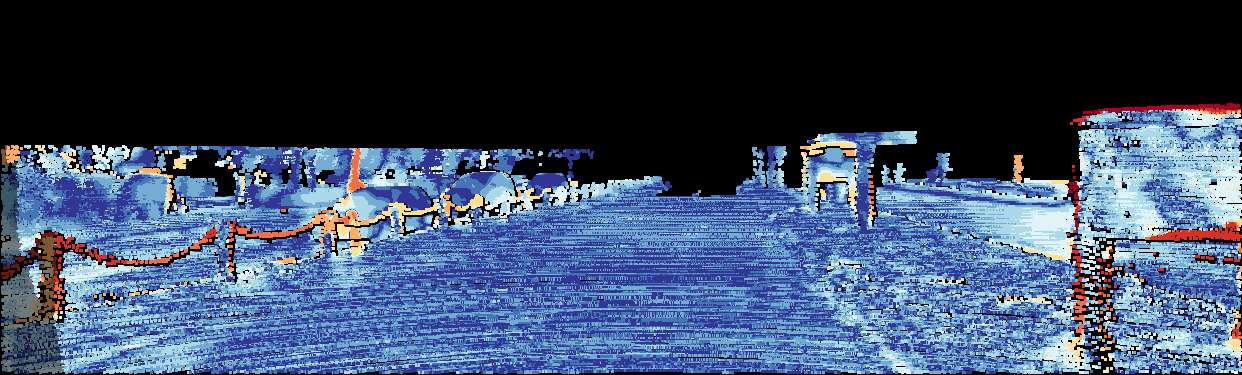} 
    \includegraphics[width=0.3\linewidth,height=0.09\linewidth]{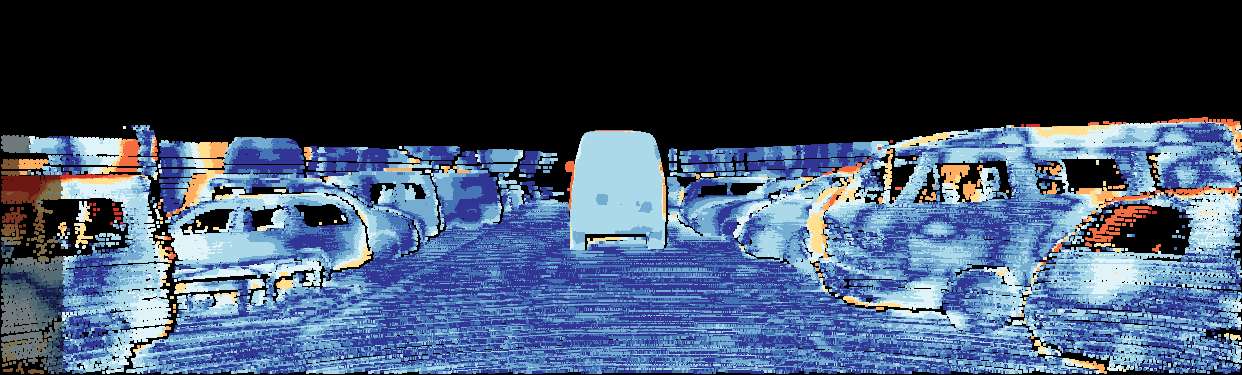}
    \includegraphics[width=0.3\linewidth,height=0.09\linewidth]{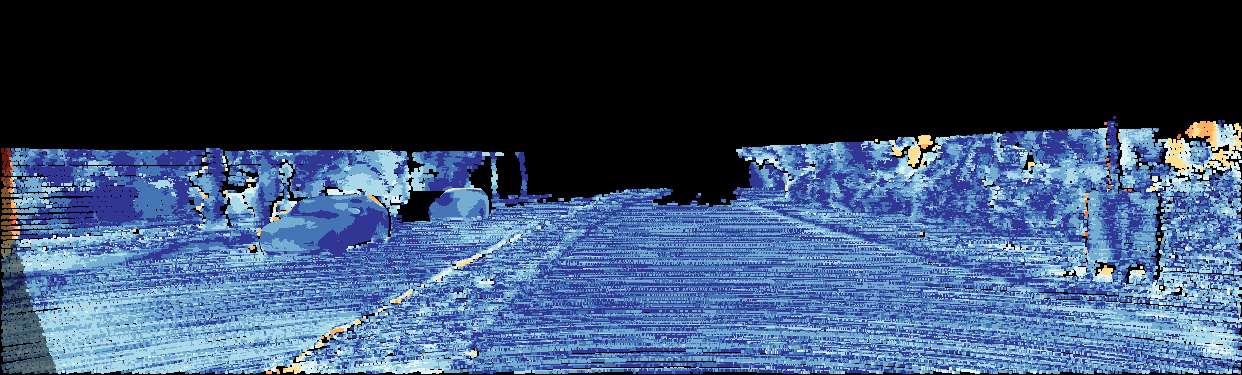}\\
    \caption{Qualitative comparison (disparity results and error maps) of aafs\cite{Chang_2020_ACCV}, StereoVAE\cite{StereoVAE}, and our \textbf{DTPnet}. Warmer colors in error maps indicate larger error.}
    \label{quality}
    \end{figure*}

\textbf{SceneFlow}: 
For a fair comparison with other state-of-the-art complex methods, the Latency in Table. \ref{table_sceneflow} is tested on Titan XP.
Despite our network being a lightweight method, we achieve competitive results compared to complex models on this dataset. Table. \ref{table_sceneflow} is arranged in descending order of end-point error (EPE), and notably, our \textbf{DPTnet} achieves the lowest EPE among all the methods.



\begin{table}[h]
    \centering
    \caption{Quantitative Comparison on Sceneflow}
    \begin{tabular}{lccccc}
    \toprule
    Model  &EPE &  Params  &FLOPs& Latency* \\
    \midrule
    PSMnet\cite{chang_pyramid_2018}      & 1.12 & 9.37 & 1083   &450 \\
    MSN3d\cite{shamsafar_mobilestereonet_2021} & 0.80& 5.22  & 578.9   & 226 \\
    DeepPruner\cite{Duggal2019ICCV}&	0.97  & 7.47 & 103.6& 182\\
    DWARF\cite{DWARF}& 1.78   & 19.6  & / &  114\\
    MSN2d\cite{shamsafar_mobilestereonet_2021} & 1.12  & 2.28 & 128.8  &107  \\
    MADnet\cite{MADnet}  & 1.66   & \textbf{0.47}  & 15.6  &  65 \\
    MABnet\cite{xing2020mabnet}  & 1.63   & 0.04  & 380.9  &  50 \\
    AAFnet\cite{Chang_2020_ACCV} & 3.90 &  0.02 &0.54 &   11\\
    \textbf{DTPnet}(ours) & \underline{1.56} & \underline{0.26}  & \underline{3.677}  & \underline{8} \\
    \bottomrule
    \multicolumn{5}{l}{\footnotesize *Latency is the single frame inference time(ms) on \textit{Titan XP}}\\
    \end{tabular}
    \label{table_sceneflow}
    \end{table}

\textbf{KITTI2015}:
Table \ref{table_kitti2015} provides a detailed comparison with other lightweight methods on the KITTI2015. 
Notably, our proposed \textbf{DTPnet} not only outperforms the other lightweight methods in terms of accuracy. We also achieve the lowest latency. (ps. the same method's inference time on Tx2 and AGX is roughly 4x times)
outperform the other methods in latency. 
The results highlight the effectiveness and competitiveness of our approach for disparity estimation on this dataset.


\subsection[short]{Qualitative results}

In Figure \ref{quality}, we present our \textbf{DTPnet} along with the results of \cite{Chang_2020_ACCV, StereoVAE}. 
A common issue observed in many methods applied to the KITTI dataset is the presence of numerous mismatches caused by low-texture areas, such as the car window. 
Our \textbf{DTPnet} learns semantic information from the context and predicts the correct disparity of the whole objects. 
\section{Conclusion}
In this paper, we proposed a real-time lightweight stereo matching network and training framework. The proposed method includes: 1. An efficient and lightweight neural network for stereo matching, 2. The Distill-then-Prune (DTP) framework for model compression. By conducting comprehensive comparisons of accuracy and FLOPs with existing methods, we demonstrate that our \textbf{DTPnet} successfully achieves our objective of providing a real-time, high-accuracy network suitable for computational resource-limited platforms. In the future, we plan to further compress the network through quantization to suit more hardware platforms while maintaining a high accuracy.

\section{Acknowledgement}
This work was supported in part by the National Natural Science Foundation of China under Grant U2013601, and the Program of Guangdong Provincial Key Laboratory of Robot Localization and Navigation Technology, under Grant 2020B121202011 and Key-Area Research and Development Program of Guangdong Province, China, under Grant 2019B010154003.




\bibliographystyle{IEEEtran}
\bibliography{include/egbib}

\end{document}